\documentclass[11pt,a4paper]{article}
\usepackage[hyperref]{acl2019}
\usepackage{times}
\usepackage{latexsym}
\aclfinalcopy 
\usepackage{times}
\usepackage{latexsym}
\usepackage{url}

\usepackage[utf8]{inputenc} 
\usepackage[english,vietnamese]{babel}
\usepackage{amsmath}
\usepackage{graphicx}
\usepackage{makecell}
\usepackage{enumitem}
\usepackage{listings}
\usepackage{footnote}
\makesavenoteenv{tabular}
\makesavenoteenv{table}
\usepackage{multirow}
\usepackage{balance}

\title{neuralVietNLP: neural NLP Tools for Vietnamese Language Processing
}

\title{Word and Char Embedding in Vietnamese: An Empirical Study and Applications.
}
\title{An Embedding Toolkit for NLP Tasks (ETNLP): a use-case in Vietnamese}
\title{OMG! What Pre-trained Word Representation Should I Use for \\a Downstream NLP Task?}
\title{ETNLP: a visual-aided systematic approach to select pre-trained embeddings for a downstream task}

\title{ETNLP: A Pipeline for Extracting, Evaluating and Visualizing Pre-trained Word Embeddings}

\title{ETNLP: a visual-aided systematic approach to select pre-trained embeddings for a downstream task}

\author{Xuan-Son Vu$^1$, Thanh Vu$^2$, Son N. Tran$^3$, Lili Jiang$^1$ \\
		$^1$Ume\r{a} University, Sweden \\
	$^2$The Australian E-Health Research Centre, CSIRO, Australia
	\\
	$^3$ The University of Tasmania, Australia;\\
	{\tt \{sonvx, lili.jiang\}@cs.umu.se}\\
	{\tt thanh.vu@csiro.au}, 
	{\tt sn.tran@utas.edu.au}
	;
}

\date{}

\begin{document}
\maketitle
\begin{abstract}
Given many recent advanced embedding models, selecting pre-trained word embedding (a.k.a., word representation) models best fit for a specific downstream task is non-trivial. 
In this paper, we propose a systematic approach, called \emph{ETNLP}, for extracting, evaluating, and visualizing multiple sets of pre-trained word embeddings to determine which embeddings should be used in a downstream task.

We demonstrate the effectiveness of the proposed approach on our pre-trained word embedding models in Vietnamese to select which models are suitable for a named entity recognition (NER) task. Specifically, we create a large Vietnamese word analogy list to evaluate and select the pre-trained embedding models for the task. We then utilize the selected embeddings for the NER task and achieve the new state-of-the-art results on the task benchmark dataset. We also apply the approach to another downstream task of privacy-guaranteed embedding selection, and show that it helps users quickly select the most suitable embeddings. In addition, we create an open-source system using the proposed systematic approach to facilitate similar studies on other NLP tasks. The source code and data are available at \url{https://github.com/vietnlp/etnlp}.
\end{abstract}


\section{Introduction}
\label{sec:1}
Word embedding, also known as word representation, represents a word as a vector capturing both syntactic and semantic information, so that the words with similar meanings should have similar vectors \cite{Levy:2014}. 
Although, the classical embedding models, such as Word2Vec~\cite{Mikolov:13}, GloVe~\cite{pennington2014glove}, fastText~\cite{bojanowski2017enriching}, have been shown to help improve the performance of existing models in a variety of Natural Language Processing (NLP) tasks like parsing~\cite{Bansal:2014}, 
topic modeling~\cite{TACL582}, 
and document classification~\cite{Taddy:2015, vu2018nihrio}. Each word is associated with a single vector leading to a challenge on using the vector across linguistic contexts~\cite{Peters:2018}. To handle the problem, recently, contextual embeddings (e.g., ELMO of~\newcite{Peters:2018}, BERT of~\newcite{devlin2018bert}) have been proposed and help existing models achieve new state-of-the-art results on many NLP tasks. Different from non-contextual embeddings, ELMO and BERT 
can capture different latent syntactic-semantic information of the same word based on its contextual uses.
Therefore, for completeness, in this paper, we incorporate both classical embeddings (i.e., Word2Vec, fastText) and contextual embeddings (i.e., ELMO, BERT) to evaluate their performances on NLP downstream tasks.

Given the fact that there are many different types of word embedding models, we argue that having a systematic pipeline to evaluate, extract, and visualize word embeddings for a downstream NLP task, is important but non-trivial. 
However, to our knowledge, there is no single comprehensive pipeline (or toolkit) which can perform all the tasks of evaluation, extraction, and visualization. For example, the recent framework called \emph{flair}~\cite{akbik2018coling} is used for training and stacking multiple embeddings but does not provide the whole pipeline of extraction, evaluation and visualization. 


In this paper, we propose \emph{ETNLP}, a systematic pipeline to extract, evaluate and visualize the pre-trained embeddings on a specific downstream NLP task (hereafter ETNLP pipeline).
The ETNLP pipeline consists of three main components which are \emph{extractor}, \emph{evaluator}, and \emph{visualizer}. Based on the vocabulary set within a downstream task, the extractor will extract a subset of word embeddings for the set to run evaluation and visualization. The results from both \emph{evaluator} and \emph{visualizer} will help researchers quickly select which embedding models should be used for the downstream NLP task. On the one hand, the \emph{evaluator} gives a concrete comparison between multiple sets of word embeddings. While, on the other hand, the \emph{visualizer} will give the sense on what type of information each set of embeddings preserves given the constraint of the vocabulary size of the downstream task.
We detail the three main components as follows.
\begin{itemize}[wide]

\item \textbf{Extractor} extracts a subset of pre-trained embeddings based on the vocabulary size of a downstream task. Moreover, given multiple sets of pre-trained embeddings, how do we get the advantage from a few or all of them? For instance, if people want to use the character embedding to handle the out-of-vocabulary (OOV) problem in Word2Vec model, they have to implement their own extractor to combine two different sets of embeddings. It is more complicated when they want to evaluate the performance of either each set of embeddings separately or the combination of the two sets. The provided \textbf{extractor} module in ETNLP will fulfill those needs seamlessly to elaborate this process in NLP applications.

\item \textbf{Evaluator} evaluates the pre-trained embeddings for a downstream task. Specifically, given multiple sets of pre-trained embeddings, how do we choose the embeddings which will potentially work best for a specific downstream task (e.g., NER)? \newcite{Mikolov:13} presented a large benchmark for embedding evaluation based on a series of analogies. However, the benchmark is only for English and there is no publicly available \emph{large} benchmark for low resource languages like Vietnamese~\cite{Vu:2014}. 
Therefore, we propose a new evaluation metric for the word analogy task in Section~\ref{sec:3}.

\item \textbf{Visualizer} visualizes the embedding space of multiple sets of word embeddings. When having a new set of word embeddings, we need to get a sense of what kinds of information (e.g., syntactic or semantic) the model does preserve. We specifically want to get samples from the embedding set to see what is the semantic similarity between different words. To fulfill this requirement, we design two different visualization strategies to explore the embedding space: (1) side-by-side visualization and (2) interactive visualization.

The side-by-side visualization helps users compare the qualities of the word similarity list between multiple embeddings (see figure~\ref{fig:side-by-side-visualizer}). It allows researchers to ``zoom-out'' and see at the overview level what is the main difference between multiple embeddings. Moreover, it can visualize large embeddings up to the memory size of the running system. Regarding implementation, we implemented this visualization from scratch running on a lightweight webserver called Flask (\url{flask.pocoo.org}). 

For the interactive visualization, it helps researchers ``zoom-in'' each embedding space to explore how each word is similar to the others. To do this, the well-known Embedding Projector (\url{projector.tensorflow.org}) is employed to explore the embedding space interactively. Unlike the side-by-side visualization, this interactive visualization can only visualize up to a certain amount of embedding vectors as long as the tensor graph is less than 2GB. This is a big limitation of the interactive visualization approach, which we plan to improve in the near future. Finally, it is worth to mention that the visualization module is dynamic and it does not require to change any codes when users want to visualize multiple pre-trained word embeddings.
\end{itemize}

To demonstrate the effectiveness of the ETNLP pipeline, we employ it to a use case in Vietnamese. Evaluating pre-trained embeddings in Vietnamese is a challenge as there is no publicly available \emph{large}\footnote{There are a couple of available datasets~\cite{N18-2032}. But the datasets are small containing only 400 words.} lexical resource similar to the word analogy list in English to evaluate the performance of pre-trained embeddings. Moreover, different from English where all word analogy records consist of a single syllable in one record (e.g., grandfather | grandmother | king | queen), in Vietnamese, there are many cases where only words formulated by multiple syllables can represent a word analogy record (e.g., ông nội | bà ngoại | vua | nữ\_hoàng). 

We propose a large word analogy list in Vietnamese which can handle the problems. Having that word analogy list constructed, we utilize different embedding models, namely Word2Vec, fastText, ELMO and BERT on Vietnamese Wikipedia data to generate different sets of word embeddings. We then utilize the word analogy list to select suitable sets of embeddings for the named entity recognition (NER) task in Vietnamese. We achieve the new state-of-the-art results on VLSP 2016\footnote{\url{http://vlsp.org.vn/vlsp2016/eval/ner}}, a Vietnamese benchmark dataset for the NER task.

Here are our key contributions in this work:
\begin{itemize}[wide]
    \item Propose a systematic pipeline (ETNLP) to evaluate, extract, and visualize multiple sets of word embeddings on a downstream task. 
    \item Release a large word analogy list in Vietnamese for evaluating multiple word embeddings. 
    \item Train and release multiple sets of word embeddings for NLP tasks in Vietnamese, wherein, their effectiveness is verified through new state-of-the-art results on a NER task in Vietnamese.
\end{itemize}

The rest of this paper is organized as follows. Section~\ref{sec:2} describes how different embedding models are trained. Section~\ref{sec:3} shows how to use ETNLP to extract, evaluate, and visualize word embeddings. Section~\ref{sec:4} explains how the word embeddings are selected for the NER task using the word analogy task. Section~\ref{sec:5} concludes the paper followed by future work.


\section{Embedding Models}
\label{sec:2}
This section details the word embedding models incorporated in our systematic pipeline.
\begin{itemize}[wide]
\item \textbf{Word2Vec (W2V)}~\cite{Mikolov:13}: a widely used method in NLP for generating word embeddings.
\item \textbf{W2V\_C2V}: the Word2Vec (W2V) model faces the OOV issue on unseen text, therefore, we provide a character2vec (C2V)~\cite{KimJSR15} embedding for unseen words. 
When the C2V is not available, it can be easily calculated from a W2V model by averaging all vectors where a character occurred. Our experiments further confirm this averaging approach is efficient.
\item \textbf{fastText}~\cite{bojanowski2016enriching}: it associates embeddings with character-based n-grams, and a word is represented as the summation of the representations of its character-based n-grams. Based on this design, fastText attempts to capture morphological information to induce word embeddings, and hence, deals better with OOV words.

\item \textbf{ELMO}~\cite{Peters:2018}: a model generates embeddings for a word based on the context it appears. Thus, we choose the contexts where the word appears in the training corpus to generate embeddings for each of its occurrences. Then the final embedding vector is the average of all its context embeddings.
\item \textbf{BERT\_\{Base, Large\}}~\cite{devlin2018bert}:  BERT makes use of Transformer, an attention mechanism that learns contextual relations between words (or sub-words) in a text. Different from ELMO, the directional models, which reads the text input sequentially (left-to-right or right-to-left), the Transformer encoder reads the entire sequence of words simultaneously. It, therefore, is considered bidirectional. This characteristic allows the model to learn the context of a word based on all of its surroundings (left and right of the word). BERT comes with two configurations called BERT\_Base (12 layers) and BERT\_Large (24 layers). To get the embedding vector of a word, we average all vectors of its subwords. Regarding contexts, similar to the ELMO model above, we choose the contexts where the word appears in the training corpus.
\end{itemize}

\section{Systematic Pipeline}
\label{sec:3}
\begin{figure}
	\centering
	\includegraphics[width=.99\linewidth]{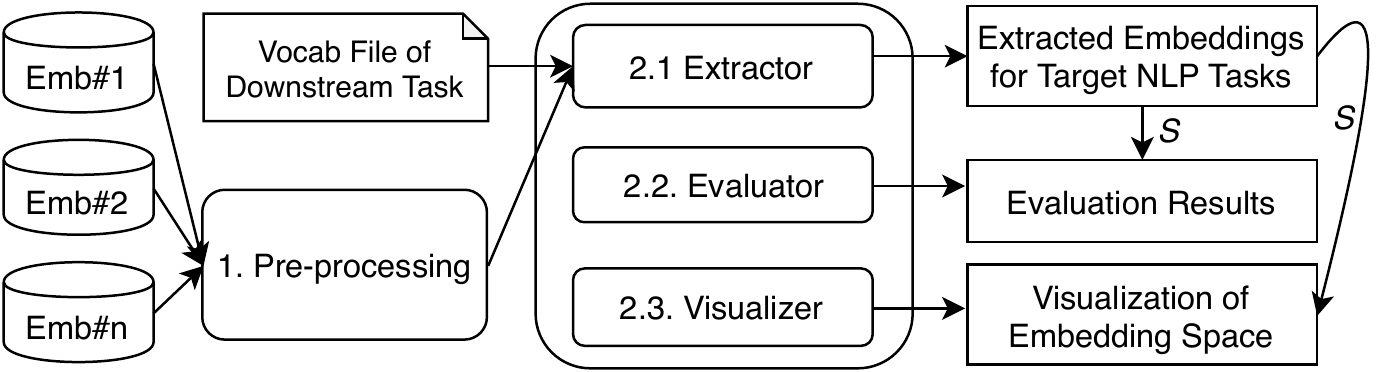}
	\caption{General process of the ETNLP pipeline where $\mathcal{S}$ is the set of extracted embeddings for Evaluation and Visualization of multiple embeddings on a downstream NLP task.}
	\label{fig:api}
\end{figure}

Figure~\ref{fig:api} shows the general process of the ETNLP pipeline. The four main processes of ETNLP are very simple to call from either the command-line or the Python API.
\begin{itemize}[wide]
    \item \textbf{Pre-processing:} since we use Word2Vec (W2V) format as the standard format for the whole process of ETNLP, we provide a pre-processing tool for converting different embedding formats to the W2V format. 


    \item \textbf{Extractor:} to extract embedding vectors at word level for the specific target NLP task (i.e., NER task in our case). For instance, the popular implementation of~\newcite{Reimers:2017:EMNLP} on the sequence tagging task
    allows users to set location for the word embeddings. The format of the file is text-based, i.e., each line contains the embedding of a word. The file then is compressed in .gz format. Figure~\ref{fig:api_extractor} shows a command-line to extract multiple embeddings for an NLP task. The argument ``-vocab'' is the location to a vocabulary list of the target NLP task (i.e., the NER task) which is extracted from the task training data.
    The option ``solveoov:1'' informs the \emph{extractor} to use Character2Vec (C2V) embedding to solve OOV words in the first embedding ``<emb\_in\#1>''. The ``-input\_c2v'' can be omitted if users wish to simply extract embeddings from the embedding list given after the ``-input\_embs'' argument. Output of this phase is a set of embeddings $\mathcal{S}$ to run on the next evaluation phase.
    
    \begin{figure}
	\centering
	\includegraphics[width=.99\linewidth]{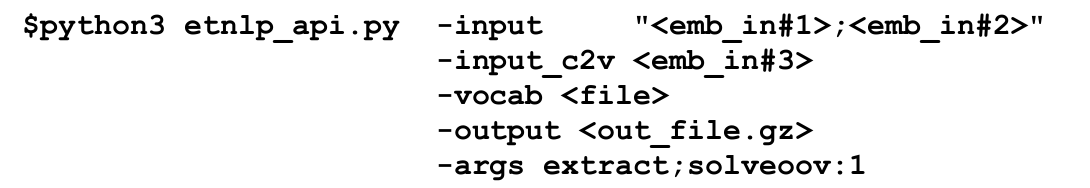}
	\vspace{-0.8cm}
	\caption{Run \emph{extractor} to export single or multiple embeddings for NLP tasks.}
	\label{fig:api_extractor}
    \end{figure}
    \vspace{-0.2cm}
    \item \textbf{Evaluator} evaluates multiple sets of embeddings (i.e., $\mathcal{S}$) on the word analogy task. Based on the performance of each set of embeddings in $\mathcal{S}$, we can decide what embeddings are used in the target NLP task. To do this evaluation, users have to set the location of the word embeddings and the word analogy list. For more convenience to represent the compound words, we use `` | '' to separate different part of a word analogy record instead of space as in the English word analogy list. Figure~\ref{fig:api_eval} shows an example of two records in the word analogy in Vietnamese (on the left) and their translation (on the right). The lower part shows a command-line to evaluate multiple sets of word embeddings on this task. Regarding this \emph{evaluator}, it is worth to note that with a huge number of possible linguistic relations (and different objectives, e.g., modeling syntactic vs. semantic properties), no embedding model is able to hold all related words close in the vector space. Therefore, only one testing schema (i.e., word analogy test) is not enough to evaluate multiple pre-trained embeddings. Thus, ETNLP is designed with the capability to be easily plugged in more tests, which makes \emph{evaluator} more robust. However, in this paper, our experimental results showed that, word analogy task is sufficient to select good embeddings for the NER task in Vietnamese.
    
    \begin{figure}
	\centering
	\includegraphics[width=.99\linewidth]{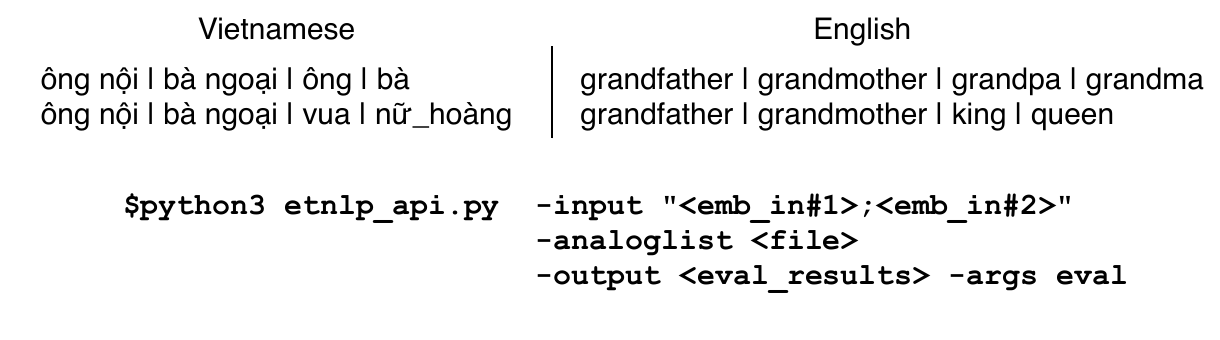}
	\vspace{-0.86cm}
	\caption{Run \emph{evaluator} on multiple word embeddings on the word analogy task.}
	\label{fig:api_eval}
    \end{figure}
\vspace{-0.25cm}

    \item \textbf{Visualizer:} to visualize given word embeddings in the argument ``-input\_embs'' in both zoom-out (the side-by-side visualization) and zoom-in (the interactive visualization) manners.
    For the zoom-out, users type a word that they want to compare the similar words in different embedding models (see Figure~\ref{fig:side-by-side-visualizer}). For the zoom-in, after the executions, embedding vectors are transformed to tensors to visualize with the Embedding Projector. Each word embedding will be set to different local port from which, users can explore the embedding space using a Web browser. Figure \ref{fig:visualizer} shows an example of the interactive visualization of ``Hà\_Nội''$_{Hanoi}$ using ELMO embeddings. See Figure~\ref{fig:api_visualizer} for an example command-line.
    
    \begin{figure}
	\centering
	\includegraphics[width=.99\linewidth]{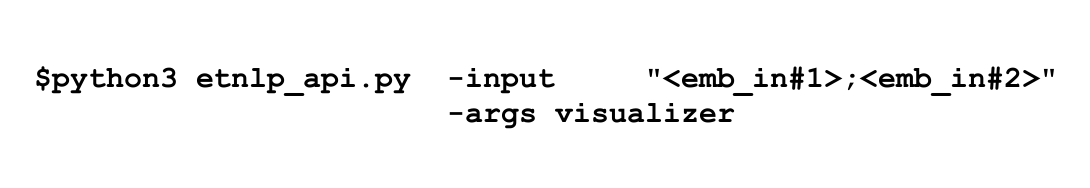}
	\vspace{-1.1cm}
	\caption{Run \emph{visualizer} to explore given pre-trained embedding models.}
	\label{fig:api_visualizer}
    \end{figure}
    
    \begin{figure*}
	\centering
	\includegraphics[width=.99\linewidth]{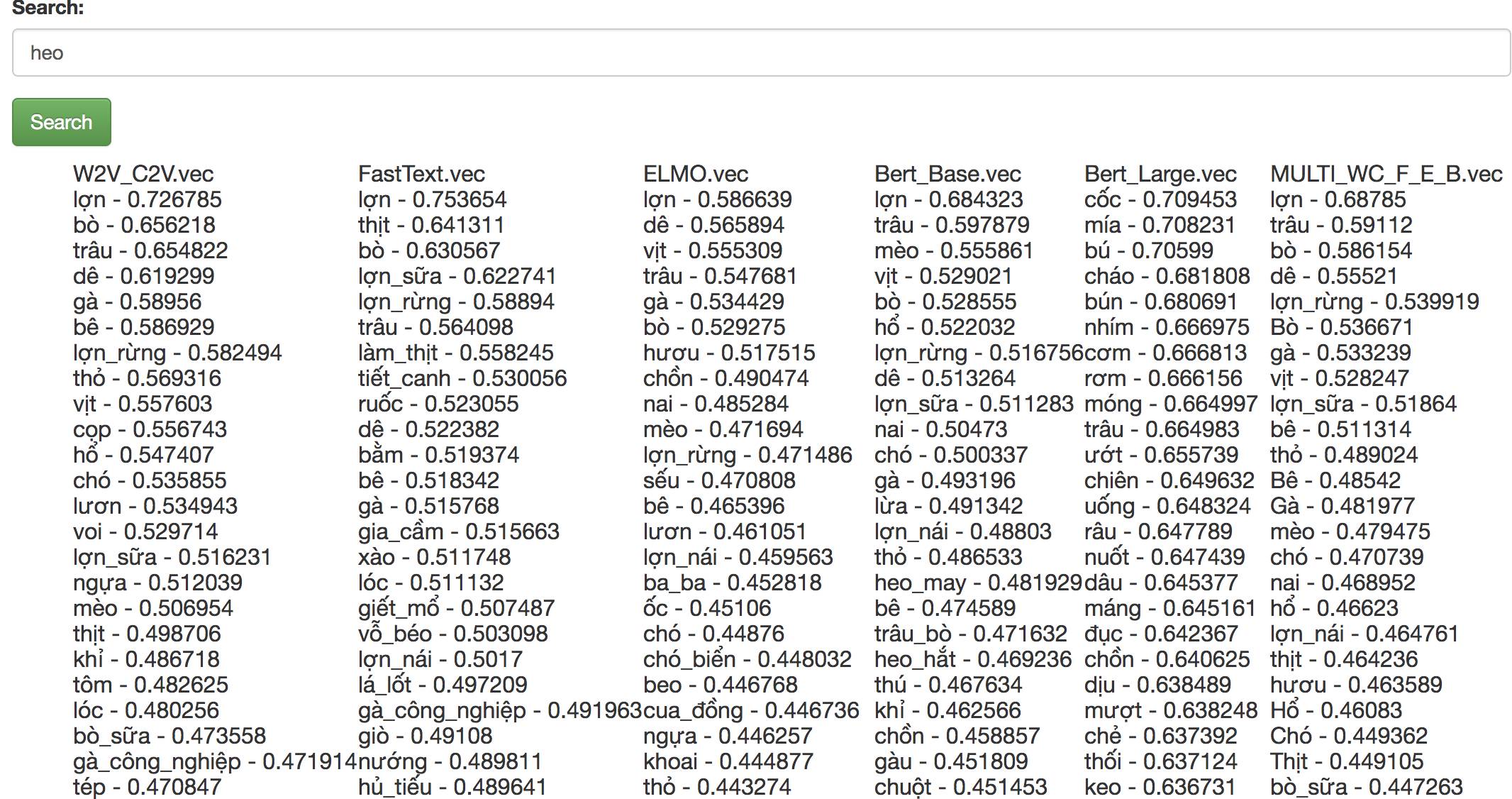}
	\caption{Side-by-side visualization for the word ``heo$_\textbf{ pig}$'' with multiple embeddings. From this visualization, we get the sense that W2V\_C2V, ELMO, and Bert\_Base mainly capture the categorical information (i.e., ``heo$_\textbf{ pig}$'' is surrounded by names of other animals, e.g., "bò$_\textbf{ cow}$", "trâu$_\textbf{ buffalo}$") while ``FastText`` captures both categorical information (i.e., surrounded by names of other animals) and related verbs to ``pig'' such as ``xào$_\textbf{ frying}$'', ``nướng$_\textbf{ grill}$''. Bert\_Large, on the other hand,  does not converge well due to the short training steps mentioned in section~\ref{sec:4}, therefore, many irrelevant words (e.g., ``cốc$_\textbf{ cup}$'', ``dịu $_\textbf{ floppy}$'') are surrounded the input word ``heo$_\textbf{ pig}$'', ``keo$_\textbf{glue}$''.}
	\label{fig:side-by-side-visualizer}
    \end{figure*}
    
    \begin{figure}
	\centering
	\includegraphics[width=.99\linewidth]{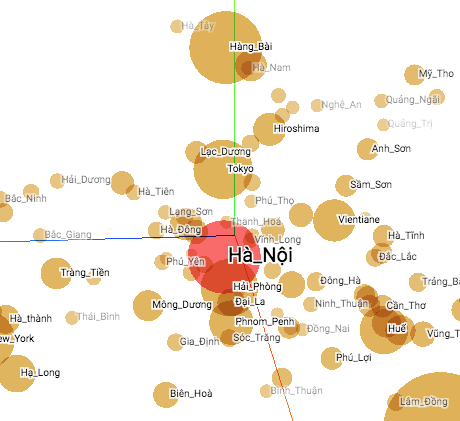}
	\vspace{-0.9cm}
	\caption{Interactive visualization for the word ``Hà\_Nội'' with ELMO embeddings where near ``Hà\_Nội'' are the names of many other cities in Vietnam (e.g., ``Hải\_Phòng$_\textbf{ Hai Phong}$'' as well as capital of other countries (e.g., Tokyo).}
	\label{fig:visualizer}
    \end{figure}
    
\end{itemize}


\section{Evaluations: a use-case in Vietnamese}
\label{sec:4}

\begin{table}[]
\caption{Evaluation results of different word embeddings on the Word Analogy Task. P-value column shows significance test results using Paired \emph{t}-tests. `*' means significant (p-value < 0.05) to the rest.}
\vspace{-0.2cm}
\centering
\scalebox{0.89}{
\begin{tabular}{l|l|l}
\hline
Model       & MAP@10 & P-value  \\ \hline
W2V\_C2V    & 0.4796 & *  \\ \hline
FastText    & 0.4970 & See [1] \& [2]  \\ \hline
ELMO        & \textbf{0.4999} & vs. FastText: 0.95
[1] \\ \hline
BERT\_Base  & 0.4609 & * \\ \hline
BERT\_Large & 0.4634 & - \\ \hline
MULTI       & 0.4906    & vs. FastText: 0.025 [2]\\ \hline
\end{tabular}
}
\label{tbl:word_analogy}
\end{table}

\begin{table*}[]
\caption{Example of five types of semantic and four (out of nine) types of syntactic questions in the word analogy list. ``NOT AVAILABLE`` means that the syntactic phenomena do not apply in Vietnamese in comparison to the list of~\newcite{Mikolov:13}.}
\vspace{-0.2cm}
\label{tbl:word_analogy_category}
\scalebox{0.8}{
\begin{tabular}{l|p{4cm}|p{6cm}|p{6cm}}
\hline
                           & Type of relationship        & Word Pair 1                          & Word Pair 2                   \\ \hline
\multirow{5}{*}{Semantic}  & capital-common-countries    & Athens | Hy\_Lạp$_\text{ Greek}$                     & | Baghdad | Irac              \\ \cline{2-4} 
                           & capital-world               & Abuja | Nigeria                      & | Thổ Nhĩ Kỳ$_\text{ Turkey}$ | Turkey         \\ \cline{2-4} 
                           & currency                    & Algeria | dinar                      & | Canada | đô la$_\text{ dollar}$              \\ \cline{2-4} 
                           & city-in-zone                & Hòa Bình$_\text{ Hoa Binh}$ | Tây Bắc Bộ$_\text{ West North}$                & | Hà Giang$_\text{ Ha Giang}$ | Đông Bắc Bộ$_\text{ East Northern}$      \\ \cline{2-4} 
                           & family                      & cậu bé$_\text{ boy}$ | cô gái$_\text{ girl}$                      & | anh trai$_\text{ brother}$ | em gái$_\text{ sister}$           \\ \hline
                           &                             &                                      &                               \\ \hline
\multirow{9}{*}{Syntactic} & gram1-adjective-to-adverb   & \multicolumn{2}{l}{NOT AVAILABLE}                                    \\ \cline{2-4} 
                           & gram2-opposite              & chấp nhận được$_\text{ acceptable}$ | không thể chấp nhận$_\text{ unacceptable}$ & | nhận thức$_\text{ aware}$ | không biết$_\text{ unaware}$      \\ \cline{2-4} 
                           & gram3-comparative           & tệ$_\text{ bad}$ | tệ hơn$_\text{ worse}$                         & | lớn$_\text{ big}$ | lớn hơn$_\text{ bigger}$               \\ \cline{2-4} 
                           & gram4-superlative           & lớn$_\text{ big}$ | lớn nhất$_\text{ biggest}$                       & | sáng$_\text{ bright}$ | sáng nhất$_\text{ brightest}$            \\ \cline{2-4} 
                           & gram5-present-participle    & \multicolumn{2}{l}{NOT AVAILABLE}                                    \\ \cline{2-4} 
                           & gram6-nationality-adjective & Albania | Tiếng Albania$_\text{ Albanian}$              & | Argentina | Tiếng Argentina$_\text{ Argentinean}$ \\ \cline{2-4} 
                           & gram7-past-tense            & \multicolumn{2}{l}{NOT AVAILABLE}                                    \\ \cline{2-4} 
                           & gram8-plural-nouns                & \multicolumn{2}{l}{NOT AVAILABLE}                                    \\ \cline{2-4} 
                           & gram9-plural-verbs          & \multicolumn{2}{l}{NOT AVAILABLE}                                    \\ \hline
\end{tabular}
}
\end{table*}

\begin{table}[]
\caption{Grid search for hyper-parameters.}
\vspace{-0.3cm}
\scalebox{0.85}{
\begin{tabular}{l|l|l|l}
\hline
Hyper-parameter          & \multicolumn{3}{l}{Search Space} \\ \hline
cemb dim (char embedding) & 50         & 100       & 500      \\ \hline
drpt (dropout rate)                   & 0.3        & 0.5       & 0.7      \\ \hline
lstm-s (LSTM size)   & 50         & 100       & 500      \\ \hline
lrate (learning rate)     & 0.0005     & 0.001     & 0.005    \\ \hline
\end{tabular}
\vspace{-0.35cm}
}
\label{tbl:search_grid}
\end{table}

\begin{table*}[ht]
\caption{Performance of the NER task using different embedding models. 
The \emph{MULTI$_{WC\_F\_E\_B}$} is the concatenation of four embeddings: W2V\_C2V, fastText, ELMO, and Bert\_Base.
``wemb dim''  is the dimension of the embedding model. VnCoreNLP\textbf{*} means we retrain the VnCoreNLP with our pre-trained embeddings.} 
\vspace{-0.3cm}
\centering
\begin{tabular}{p{5.5cm}|l|l|l|l|l|l}
\hline
 & F1 & wemb dim & cemb dim & drpt & lstm-s & lrate  \\ \hline
BiLC3 ~\cite{Ma:2016} & 88.28  & 300   & -   &  -  &  -  & -   \\\hline
VNER~\cite{Ngan:2018}      & 89.58   &  300   & 300  & 0.6   & -   & 0.001  \\ \hline
VnCoreNLP~\cite{VU:2018}       & 88.55   &  300   &  -  & -   & -   & -  \\ \hline
VnCoreNLP $^{(\textbf{*})}$       & \textbf{91.30}   & 1024   &  -  & -   & -   & -  \\ \hline
BiLC3 + W2V  & \emph{89.01}   & 300     & 50   & 0.5  & 100   & 0.0005 \\ \hline
BiLC3 + BERT-Base   & \emph{88.26}   & 768  & 500  & 0.3  & 100   & 0.0005 \\ \hline
BiLC3 + W2V\_C2V   & \emph{89.46}    & 300   & 100  & 0.5  & 500   & 0.0005 \\ \hline
BiLC3 + fastText  & 89.65     & 300    & 500  & 0.3  & 100   & 0.001  \\ \hline
BiLC3 + ELMO   & 89.67   & 1024    & 100  & 0.7  & 500   & 0.0005 \\ \hline
BiLC3 + MULTI$_{WC\_F\_E\_B}$   & \textbf{91.09}  & 2392  & 100  & 0.7  & 100   & 0.001 \\  \hline
\end{tabular}
\
\label{tab:ner_results_normal}
\end{table*}
\vspace{-0.3cm}

 \begin{figure*}
	\centering
	\includegraphics[width=.99\linewidth,height=6.3cm]{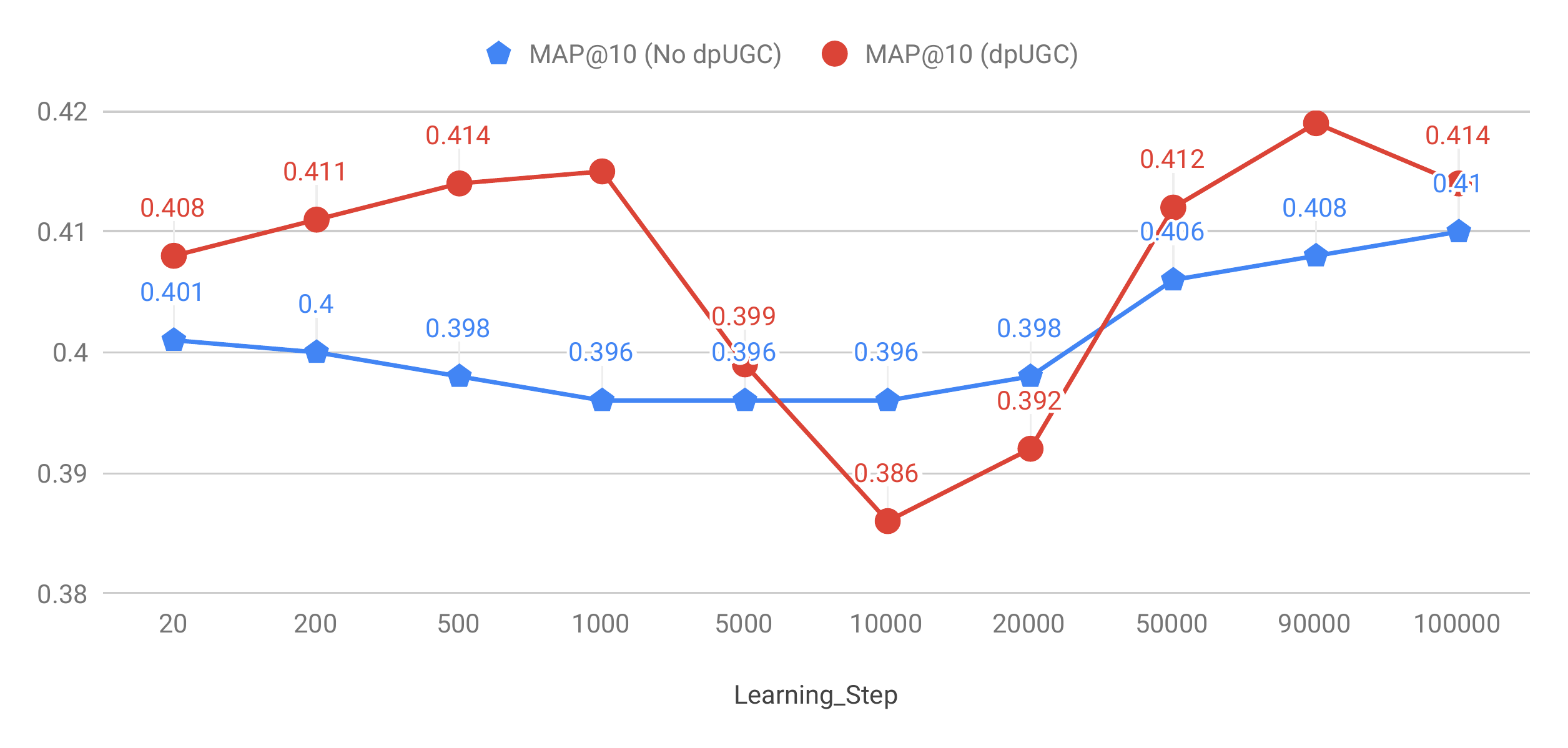}
	\vspace{-0.6cm}
	\caption{Evaluation results of different word embeddings trained using \textbf{dpUGC} and \textbf{No dpUGC} (i.e., one option in dpUGC to train embeddings without privacy guarantee for comparison) on the Word Analogy Task.}
	\label{fig:dpugcAnalogy}
\end{figure*}

\begin{table*}[]
\caption{P-values of the paired t-tests between embeddings obtained using dpUGC at different learning step (Emb@L). ``-'' denotes values of these entries in the upper triangular matrix are the values of the transposed entries in the lower triangular matrix. P-values in \textbf{bold font} are statistical significance at the level of $0.05$.}
\vspace{-0.2cm}
\scalebox{0.95}{
\begin{tabular}{|l|l|l|l|l|l|l|l|l|l|p{0.6cm}|}
\hline
Emb@L & 20               & 200               & 500     & 1000    & 5000            & 10000   & 20000            & 50000  & 90K  & 100K \\ \hline
20           & 1                &     -              &    -     &    -     &     -            &   -      &       -           &   -    &   -     & -       \\ \hline
200          & 0.0578           & 1                 &      -   &   -      &     -            &   -      &       -           &   -     &  -      & -       \\ \hline
500          & \textbf{0.0074}  & 0.1809           & 1       &    -     &      -           &    -     &      -            &  -      &    -    &   -     \\ \hline
1000         & \textbf{0.0053}  & 0.169             & 0.9031 & 1       &     -            &    -     &        -          &    -    &      -  &    -    \\ \hline
5000         & \textbf{0.0178} & \textbf{0.0009}   & 6.992   & 1.6242  & 1               &     -    &         -         &    -    &    -    &   -     \\ \hline
10000        & 2.543           & 6.9872            & 2.25867 & 9.3987  & \textbf{0.001}         & 1       &       -           &    -    &   -     &   -     \\ \hline
20000        & \textbf{0.0016}  & \textbf{0.0001} & 1.757   & 9.6053  & 0.112           & 0.1819  & 1                &   -     &    -    &  -      \\ \hline
50000        & 0.5077           & 0.9023           & 0.73137 & 0.7003 & \textbf{0.031}  & 5.0673 & \textbf{0.0001} & 1      &    -    &   -     \\ \hline
90K        & 0.1205           & 0.2878            & 0.5127 & 0.5323  & \textbf{0.0049} & 2.4211  & \textbf{0.0001} & 0.2688 & 1      &  -      \\ \hline
100K       & 0.3777           & 0.6822            & 0.9932  & 0.9764  & \textbf{0.0357} & 8.2638  & \textbf{0.0019} & 0.7274 & 0.2758 & 1      \\ \hline
\end{tabular}
}
\label{tbl:dpugc_pvalues}
\end{table*}

\subsection{Training word embeddings}
\label{sec.4.1}
We trained embedding models detailed in Section \ref{sec:2} on the Wikipedia dump in Vietnamese\footnote{\url{https://goo.gl/8WNfyZ}}. We then apply sentence tokenization and word segmentation provided by VnCoreNLP~\cite{VU:2018, NGUYEN18.55} to pre-process all documents. It is noted that, for BERT model, we have to (1) format the data differently for the next sentence prediction task; and (2) use SentencePiece~\cite{Kudo:2018} to tokenize the data for learning the pre-trained embedding. It is worth noting that due to the limitation in computing resources, we can only run BERT\_Base for 900,000 update steps and BERT\_Large for 60,000 update steps. We, therefore, do not report the result of BERT\_Large for a fair comparison. We also create \emph{MULTI} embeddings by concatenating four sets of embeddings (i.e., W2V\_C2V, fastText, ELMO and BERT\_Base)~\footnote{We do not use W2V here because W2V\_C2V is W2V with the use of character embedding to deal with OOV.}.

\subsection{Dataset}
\label{dataset}
The named entity recognition (NER) shared task at the 2016 VLSP workshop provides a dataset of 16,861 manually annotated sentences for training and development, and a set of 2,831 manually annotated sentences for test, with four NER labels PER, LOC, ORG, and MISC. The data was published in 2016 and recently reported in~\newcite{Nguyen:19}. It is a standard benchmark on the NER task and has been used in~\cite{VU:2018,Ngan:2018}.
It is noted that, in the original dataset, each word representing a full personal name are
separated into syllables that constitute the word. Because this annotation scheme results in an unrealistic scenario for a pipeline evaluation~\cite{VU:2018}, therefore, we tested on a ``modified'' VLSP 2016 corpus where we merge contiguous syllables constituting a full name to form a word. This similar setup was also used in~\cite{VU:2018,Ngan:2018}, the current state-of-the-art approaches.

\subsection{Word Analogy Task}
To measure the quality of different sets of embeddings in Vietnamese, similar to~\newcite{Mikolov:13}, we define a word analogy list consisting of 9,802 word analogy records. To create the list, we selected suitable categories from the English word analogy list and then translated them to Vietnamese. We also added customized categories which are suitable for Vietnamese (e.g., cities and their zones in Vietnam). 
Different from~\cite{Mikolov:13}, five categories: ``Adjective to adverb'', ``Present Participle'', ``Past tense'', ``Plural nouns'', ``Plural verbs'' were not used to be translated in Vietnamese since the same syntactic phenomena does not exist in Vietnamese. Table~\ref{tbl:word_analogy_category} shows the list of categories and their examples of the constructed word analogy list in Vietnamese. Since most of this process is automatically done, it can be applied easily to other languages.
To know which set of word embeddings potentially works better for a target downstream task,
we limit the vocabulary of the embeddings similar to vocabulary of the task (i.e., the NER task).
Thus, only 3,135 word analogy records are being evaluated for the NER dataset (Section \ref{dataset}).

Regarding the evaluation metric,~\newcite{Mikolov:13} used accuracy metric to measure the quality of word embeddings on the task in which only when the expected word is on top of the prediction list, then the model gets +1 for true positive count. However, this is not a well-suited metric in low resource languages where training corpus is relatively small, i.e., 233M tokens in Vietnamese Wiki compared to 6B tokens in Google News corpus. Therefore, we change to use mean average precision (MAP) metric to measure quality of the word analogy task. 
MAP is widely used in information retrieval to evaluate results based on the top\emph{K} returned results~\cite{Vu:CiCLing:2019b}. We use MAP@10 in this paper. 
Table~\ref{tbl:word_analogy} shows evaluation results of different sets of embeddings on the word analogy task. The \emph{evaluator} of ETNLP also shows P-value using the paired t-tests on the raw MAP@10 scores (i.e., before averaging) between different sets of embeddings. The P-values (Table~\ref{tbl:word_analogy}) show that the performances of the top three sets of word embeddings  (i.e., fastText, ELMO, and MULTI), are significantly better than the remainders but there is no significant difference between the three. Therefore, these sets of embeddings will be selected for NER task.
\subsection{NER Task}
\vspace{-0.1cm}
\textbf{Model:} We apply the current most well-known neural network architecture for NER task of~\newcite{Ma:2016} with no modification in its architecture, namely, \emph{BiLSTM-CRF+CNN-char (BiLC3)}. Only in the embedding layer, a different set of word embeddings is used to evaluate their effectiveness. Regarding experiments, we perform a grid search for hyper-parameters and select the best parameters on the validation set to run on the test set. Table~\ref{tbl:search_grid} presents the value ranges we used to search for the best hyper-parameters. We also follow the same setting as in ~\cite{VU:2018} to use the \emph{last} 2000 records in the training data as the validation set. Moreover, due to the availability of the VnCoreNLP code, we also retrain their model with our pre-trained embeddings (\emph{VnCoreNLP$^{*}$}).

\textbf{Main results:} Table~\ref{tab:ner_results_normal} shows the results of NER task using different word embeddings. It clearly shows that, by using the pre-trained embeddings on Vietnamese Wikipedia data, we can achieve the new state-of-the-art results on the task. The reason might be that FastText, ELMO and MULTI can handle OOV words as well as capture better the context of the words. Moreover, learning the embeddings from a formal dataset like Wikipedia is beneficial for the NER task. This also verified the fact that using our pre-trained embeddings on VnCoreNLP helps significantly boost its performance. Table~\ref{tab:ner_results_normal} also shows the F1 scores of W2V, W2V\_C2V and BERT\_Base embeddings which are worse than three \emph{selected} embeddings (i.e., fastText, ELMO and MULTI). This might indicate that using word analogy to select embeddings for downstream NLP tasks is sensible. 








\subsection{Privacy-Guaranteed Embedding Selection Task}
In this section, we show how to apply ETNLP to another downstream task of privacy-guaranteed embedding selection. \newcite{Vu:CiCLing:2019b} introduced dpUGC to guarantee privacy for word embeddings. The main intuition behind dpUGC is that, when the embedding is trained on very sensitive text corpus (e.g., medical text data), it has to guarantee privacy at the highest level to prevent privacy leakage. However, among many embeddings at different learning steps of dpUGC, how to choose a suitable embedding to achieve a good trade-off between data privacy and data utility is a key challenge. To this end, we propose to apply ETNLP into this scenario to select good embeddings for knowledge sharing using dpUGC. 

Similar to \newcite{Vu:CiCLing:2019b}, we trained 20 different embeddings from 10 different learning steps while training on the same Vietnamese Wikipedia dataset as used in Section \ref{sec.4.1} with (dpUGC) and without privacy-guarantee (No dpUGC) to evaluate their performances.
Figure~\ref{fig:dpugcAnalogy} shows that the pre-trained embedding at learning\_step 1000 (Emb@1000) seems to be a good word embbedding candidate to have a good trade-off between privacy guarantee and data utility. Emb@1000 was in favor because of two reasons. Firstly, in training privacy-guaranteed embeddings, we try to stop as early as possible since the more training steps we run, the higher privacy we have to sacrifice~\cite{Vu:CiCLing:2019b}. Secondly, its performance in the Word Analogy Task was more or less similar to the other good embedding at the learning step 90K (i.e., Emb@90K). In fact, from Table~\ref{tbl:dpugc_pvalues} we know that the performance between Emb@1000 and Emb@90K learning steps are not significant difference. Therefore, selecting the pre-trained embedding at the learning step 1000 is the best option for privacy-guaranteed embedding using dpUGC. 
In summary, in this task, we showed how ETNLP can be used to select a good word embedding candidate for privacy-guaranteed knowledge sharing. Normally, this selection process is very time consuming, however, it is much easier with ETNLP since it allows users to import multiple embeddings for running evaluations.

\section{Conclusions}
\label{sec:5}
We have presented a new systematic pipeline, ETNLP, for extracting, evaluating and visualizing multiple pre-trained embeddings on a specific downstream task. The ETNLP pipeline was designed with three principles in mind: (1) easy to apply on any language processing task, (2) better performance, and (3) be able to handle unknown vocabulary in real-world data (i.e., using C2V (char to vec)). The evaluation of the approach in (1) Vietnamese NER task and (2) privacy-guaranteed embedding selection task showed its effectiveness. 

In the future, we plan to support more embeddings in different languages, especially in low resource languages. We will also support new ways to explore the embedding spaces including at phrase and subword levels.

\section*{Acknowledgement} {
This work is supported by the Federated Database project funded by Ume\r{a} University, Sweden.
The computations were performed on resources
provided by the Swedish National Infrastructure for Computing (SNIC)
at HPC2N.
}

\bibliographystyle{aclnatbib}
\balance
\bibliography{ranlp2019_1,ranlp2019_dp}

\end{document}